\documentclass[runningheads]{llncs}
\usepackage[T1]{fontenc}
\usepackage{hyperref}
\usepackage{graphicx,verbatim}
\usepackage{amsmath}
\usepackage{amssymb}
\usepackage{booktabs}
\usepackage{multirow}
\usepackage{makecell}
\usepackage{siunitx}
\usepackage{threeparttable}
\usepackage{colortbl}
\usepackage{placeins}

\usepackage[table]{xcolor}
\usepackage{colortbl}

\definecolor{pmGreen}{RGB}{0,140,0}
\definecolor{oursBlue}{RGB}{220,235,255}
\definecolor{teacherGray}{RGB}{120,120,120}

\newcommand{\pmg}[1]{{\scriptsize\textcolor{pmGreen}{$\pm$#1}}}

\begin{document}
\title{Detail Consistent Stage-Wise Distillation for Efficient 3D MRI Segmentation}
%

\author{Mengchen Fan\inst{1} \and
Baocheng Geng\inst{1} \and
Xi Xiao\inst{1} \and
Tianyang Wang\inst{1} \and
Siyuan Mei\inst{2} \and
Pulin Che\inst{1} \and
Xiaoqian Jiang\inst{3}\textsuperscript{$\star$} \and
Qizhen Lan\inst{3}\thanks{Corresponding author.}}


%
\authorrunning{M. Fan et al.}
%

\institute{University of Alabama at Birmingham, Birmingham, AL, USA\\
\email{\{fanm,bgeng,xxiao,tw2,pche\}@uab.edu}
\and
Friedrich-Alexander-Universit\"at Erlangen-N\"urnberg, Erlangen, Germany\\
\email{siyuan.mei@fau.de}
\and
UTHealth Houston, Houston, TX, USA\\
\email{\{xiaoqian.jiang,qizhen.lan\}@uth.tmc.edu}}
  
\maketitle              
\begin{abstract}
Deploying high-performing 3D medical image segmenters (e.g., nnU-Net) is often limited by memory footprint and inference latency. Compression is therefore necessary, but compact 3D encoders tend to lose fine structural cues (small lesions and sharp boundaries) as downsampling repeats across multi-resolution stages. We propose \emph{Detail Consistent Distillation (DCD)}, a stage-wise distillation framework that preserves structural detail across scales by aligning teacher--student features in a wavelet-decomposed representation. At each encoder stage, DCD distills directional detail components in the wavelet domain while leaving the coarse approximation comparatively unconstrained, avoiding over-regularization of global semantics. DCD is used only during training and introduces no inference-time overhead. Experiments on the BraTS 2024 and ISLES 2022 benchmarks demonstrate that our approach achieves superior performance in MRI segmentation using 3D multi-modal data. Code and implementation details for DCD are publicly available at
\url{https://github.com/ClinicaAlpha/DCD-3D-MedSeg}.


\end{abstract}
\section{Introduction}

Accurate 3D lesion segmentation in brain MRI supports quantitative neuroimaging and time-sensitive workflows, including tumor sub-region assessment and ischemic stroke lesion delineation \cite{Menze2015BraTS,hernandez2022isles}. Encoder--decoder U-Net designs \cite{Ronneberger2015UNet,Cicek2016UNet3D} and standardized training pipelines such as nnU-Net \cite{Isensee2021nnUNet} have made strong 3D baselines widely accessible, but volumetric models remain expensive in memory and latency, limiting deployment on constrained hardware and in high-throughput settings \cite{Wang2022MemoryEfficient3D}, where hardware-aware model design has become increasingly important \cite{chen2025autoneuralcodesigningvisionlanguagemodels,fan2025pfeddst,fan2025unified,fan2025prune}.

Model compression is therefore necessary. Knowledge Distillation (KD) transfers behavior from a strong teacher to a compact student during training \cite{Bucila2006ModelCompression,hinton2015distilling} and has been used to compress segmentation ensembles into smaller networks while retaining accuracy \cite{lan2025acam,lan2026reco}. However, compact 3D segmenters often fail in a characteristic way: they preserve coarse localization but lose fine boundaries and small or thin structures. We attribute this to progressive suppression of detail-sensitive representations as capacity shrinks and downsampling repeats over the encoder hierarchy.
\begin{figure}[t]
  \centering
  \includegraphics[width=\linewidth]{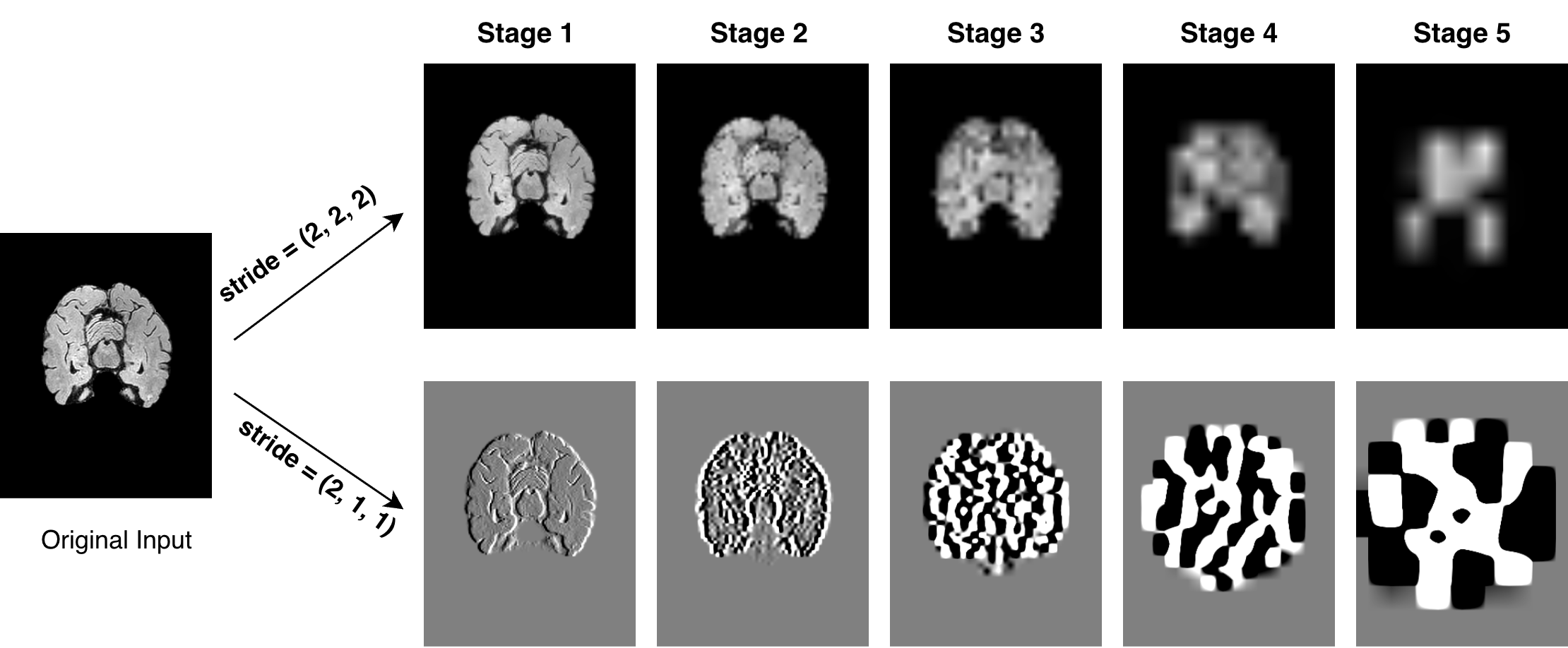}
  \caption{Five successive downsampling stages (stride=2 in $D,H,W$) shown for two complementary responses. The intensity-like stream (top row) progressively loses fine anatomical details, while the derivative/edge response (bottom row, $D_x$) rapidly degrades into coarse, aliased patterns, illustrating joint loss of global structure and local detail under repeated decimation.}
  \label{fig:downsampling}
\end{figure}

Two effects motivate explicit supervision of detail components in compressed 3D UNets. First, strided downsampling is a lossy sampling operation (as shown in Fig. \ref{fig:downsampling}). It can discard or alias high-frequency content with errors compounding across stages \cite{Zhang2019ShiftInvariant}. Second, deep networks exhibit spectral bias: they fit lower-frequency components earlier and more readily, while higher-frequency variation is harder to represent and typically requires more capacity \cite{Rahaman2019SpectralBias,Xu2020FrequencyPrinciple}. Together, these effects make conventional ``match everything" distillation prone to producing over-smoothed students even when the teacher’s predictions are sharp.

Frequency-domain supervision provides a direct handle on the components most vulnerable to over-smoothing. The discrete wavelet transform (DWT) decomposes a volume into a low-frequency
approximation and directional detail subbands with exact reconstruction, and wavelets are established
tools for separating scale-specific information in biomedical signals and images. \cite{Mallat1989Wavelet,Daubechies1992TenLectures,UnserAldroubi1996WaveletsBiomedical}. In MRI, this perspective is physically grounded by $k$-space: low spatial frequencies dominate global contrast, while
higher spatial frequencies contribute disproportionately to edges and fine detail \cite{moratal2008k,Mezrich1995KSpace}. However, not all high frequency
content is equally informative; high frequency bands can also capture acquisition noise and
scanner-specific artifacts, and MRI statistics differ from natural scenes \cite{Xu2019MRIStats}. This motivates selective frequency supervision that emphasizes boundary relevant detail while avoiding the noise dominated components.

Recent KD work has explored frequency-aware distillation in 2D natural image settings \cite{lan2026performance,Zhang2022Wavelet,Zhang2024FreeKD}, but medical images have different spectral statistics than natural scenes \cite{Field1987NaturalImages,Ruderman1994ScalingWoods,Xu2019MRIStats}, so indiscriminate frequency matching can align noise or acquisition-specific artifacts rather than clinically meaningful structure.

Motivated by this, we propose \textbf{Detail Consistent Stage-Wise Distillation (DCD)} for efficient 3D medical image segmentation. DCD aligns teacher and student encoder features at each resolution stage by applying a
3D DWT to isolate wavelet subbands and distilling only directional detail components, while keeping the low frequency approximation comparatively unconstrained to preserve global semantics.
To maintain spatially interpretable supervision, we reconstruct detail-only features via inverse DWT (IDWT) before
performing alignment, yielding a spatial-domain loss that targets detail content. DCD is used only during training and introduces no inference-time overhead. We evaluate on brain tumor and stroke lesion
segmentation benchmarks \cite{Menze2015BraTS,hernandez2022isles} and report improved accuracy–efficiency trade-offs under nnU-Net compression. Our main contributions are:
\begin{enumerate}
    \item A wavelet-domain subband selection strategy that distills directional detail while excluding the most noise-prone extreme high-frequency band;
    \item A stage-wise distillation framework that explicitly targets detail subspaces in compressed 3D U-Net architectures; and
    \item Empirical evaluation on 3D
    brain tumor and stroke lesion benchmarks showing improved boundary-sensitive quality under compression, without adding inference-time cost.
\end{enumerate}

\section{Proposed Method}
\begin{figure}[t]
  \centering
  \includegraphics[width=\linewidth]{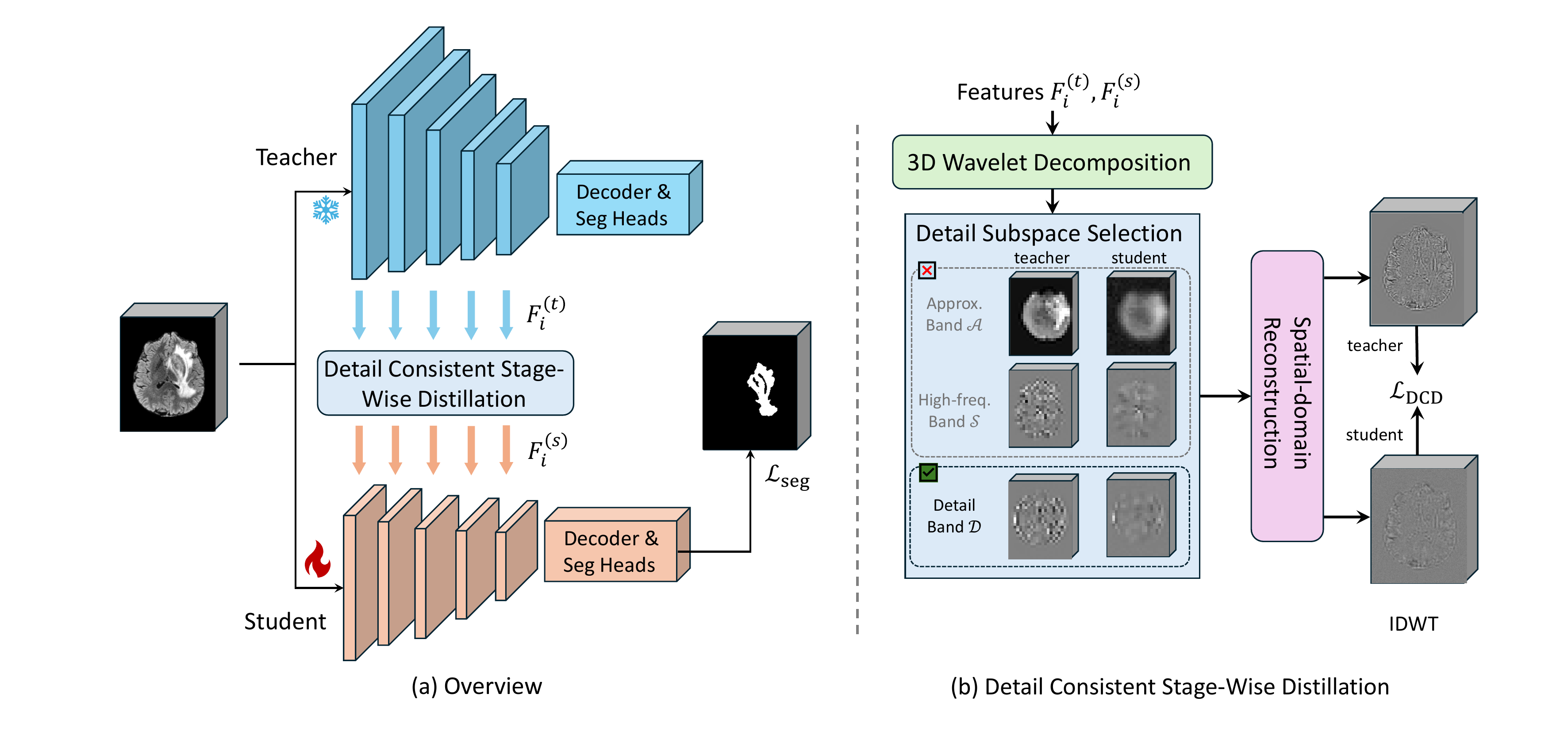}
  \caption{Overview of Detail Consistent Distillation (DCD), (a) Teacher–student distillation in a U-Net-like architecture, where DCD is applied at each encoder stage and optimized jointly with the segmentation loss. (b) Stage-wise DCD. Given stage features, we perform a 3D wavelet decomposition, select the detail subspace (band $\mathcal{D}$), reconstruct detail only features via IDWT, and compute $\mathcal{L}_{\mathrm{DCD}}$  as the MSE between teacher and student.}
  \label{fig:overview}
\end{figure}

We distill a compact student model from a high-capacity teacher (as shown in Fig.~\ref{fig:overview}) by explicitly emphasizing
detail bearing responses that tend to degrade under compression and repeated downsampling.
Given an input volume $x$, let $F_i^{(t)} \in \mathbb{R}^{C_i^{(t)} \times H_i \times W_i \times D_i}$ and
$F_i^{(s)} \in \mathbb{R}^{C_i^{(s)} \times H_i \times W_i \times D_i}$ denote encoder stage features of the teacher and student at stage $i\in\{1,\dots,M\}$. The student is obtained by uniform channel reduction, i.e., $C^{(s)} = C^{(t)}/{r}$ for a reduction factor $r$. To align channel dimensions for distillation, we use a learnable $1\times1\times1$ projection $\phi_i(\cdot)$. The teacher parameters are frozen during distillation.

\subsection{Capacity-Induced Spectral Bias}
\label{sec:2.1}

When model capacity is reduced, the student tends to preserve components that dominate the loss
and feature energy, while fine structural variations are more easily suppressed.
This behavior is consistent with the spectral bias of neural networks. During gradient based training, lower frequency components are often learned more readily than higher frequency variation, which can
require greater representational capacity \cite{Rahaman2019SpectralBias,Xu2020FrequencyPrinciple,xu2025overview,fan2024interpretable}. 

Therefore, KD in U-Net-like architectures inherently exhibits
a spectral bias. In U-Net-like architectures that repeatedly
downsample along the encoder hierarchy, this bias can manifest as students that match coarse semantics while under representing boundary critical detail. 

We summarize the above mismatch as the following motivating inequality between teacher and student features at stage $i$:
\begin{equation}
\label{eq:spectral_imbalance}
\left\|\mathcal{P}_{\mathrm{hi}}\!\left(F_i^{(t)}\right)-\mathcal{P}_{\mathrm{hi}}\!\left(F_i^{(s)}\right)\right\|_2
\;>\;
\left\|\mathcal{P}_{\mathrm{lo}}\!\left(F_i^{(t)}\right)-\mathcal{P}_{\mathrm{lo}}\!\left(F_i^{(s)}\right)\right\|_2,
\end{equation}
where $\mathcal{P}_{\mathrm{hi}}(\cdot)$ and $\mathcal{P}_{\mathrm{lo}}(\cdot)$ denote complementary projections onto high- and low-frequency components.

\subsection{Wavelet Detail Subspace Selection}

\textbf{Wavelet Subspace Decomposition} To obtain a structured multi-resolution decomposition, we apply a 3D Discrete Wavelet Transform (DWT)
independently to each channel of a stage feature tensor $F \in \mathbb{R}^{C \times H \times W \times D}$:
\begin{equation}
\label{eq:dwt_def}
\left\{F^{(\beta)}_l\right\}_{\beta \in \{L,H\}^3} = \xi_l (F), \qquad
 \beta \in \{LLL,LLH,\dots,HHH\},
\end{equation}
We
use an $l$-level decomposition by recursively applying the transform to the approximation subband LLL, $\xi_l (\cdot)$ is the level $l$ 3D-DWT and $\beta$ indexes low-/high-pass filtering along the three spatial axes. In this paper, we set $l=3$ for all distillation experiments.

For distillation, we partition the subbands as
\begin{equation}
\label{eq:subband_partition}
\mathcal{A}=\{LLL\}, \qquad
\mathcal{S}=\{HHH\}, \qquad
\mathcal{D}=\{L,H\}^3\setminus\{LLL,HHH\}.
\end{equation}
$\mathcal{A}$ captures coarse structure, while $\mathcal{D}$ contains detail components
that are strongly coupled to boundaries and thin structures, and $\mathcal{S}$ represents simultaneously high-pass filtering along all three axes
and is often more noise-sensitive in MRI.

\begin{proposition}[Motivation for excluding $\mathcal{S}$ under a simplified MRI noise model]
\label{prop}
Assume a stage feature can be decomposed as $F = F_{\mathrm{sig}} + \epsilon$, where $\epsilon$ is additive, zero-mean, i.i.d.\ noise
with variance $\sigma^2$ per voxel (a standard white-noise approximation for high-frequency noise components). $\xi(\cdot)$ is orthonormal.
Then energy is preserved and the expected noise energy per subband
satisfies
\begin{equation}
\label{eq:4}
\|\xi(\epsilon)\|_2^2 = \|\epsilon\|_2^2,
\qquad
\mathbb{E}\!\left[\|\xi(\epsilon)^{(\beta)}\|_2^2\right] = \sigma^2 N_\beta,
\end{equation}
where $N_\beta$ is the number of coefficients in subband $\beta$. In MRI, signal energy often decays with spatial
frequency (k-space center dominance), so the expected signal energy in the band
$\mathcal{S}=\{\mathrm{HHH}\}$ can be substantially smaller than that in the directional detail bands $\beta \in \mathcal{D}$,
\begin{equation}
\mathbb{E}\!\left[\|\xi(F_{\mathrm{sig}})^{(\mathcal{S})}\|_2^2\right]
\ll
\max_{\beta \in \mathcal{D}}
\mathbb{E}\!\left[\|\xi(F_{\mathrm{sig}})^{(\beta)}\|_2^2\right].
\label{eq:5}
\end{equation}

Defining the subband SNR as
\(
\mathrm{SNR}_\beta
=
\frac{\mathbb{E}\!\left[\|\xi(F_{\mathrm{sig}})^{(\beta)}\|_2^2\right]}
{\sigma^2 N_\beta}
\), which implies $\mathrm{SNR}_{\mathcal{S}}$ is typically much smaller than $\mathrm{SNR}_\beta$ for $\beta \in \mathcal{D}$ under the condition in Eq.~\ref{eq:5}. This motivates excluding $\mathcal{S}$ from distillation to avoid encouraging the student to learn noise dominated components.
\end{proposition}
\noindent \textbf{Subband $\mathcal{D}$ selection}
Full spectrum feature matching is typically dominated by high energy low-frequency components, which
weakens supervision on detail-bearing bands and amplifies the capacity-induced mismatch in Eq.~\eqref{eq:spectral_imbalance}.
DCD therefore removes the distillation constraint on the approximation band $\mathcal{A}$, but we do not remove low frequency
features from the network, so the student is not over-regularized on global semantics. Moreover, by Proposition~\ref{prop}, the band $\mathcal{S}$ tends to have lower effective SNR and can inject noise into the distillation signal. Therefore, we choose $\mathcal{D}$ as the distillation frequency subspace to preserve directional details while avoiding both low-frequency dominance $\mathcal{A}$ and noise-prone components $\mathcal{S}$.

\subsection{Detail Consistent Stage-Wise Distillation}
We define a detail projection that reconstructs a spatial-domain representation using only the detail subbands $\mathcal{D}$:
\begin{equation}
\label{eq:detail_projection}
\mathcal{P}_{\mathcal{D}}(F)
=
\xi_l^{-1}\!\left(\left\{\widetilde{F}^{(\beta)}_l\right\}\right),
\qquad
\widetilde{F}_l^{(\beta)}=
\begin{cases}
F_l^{(\beta)}, & \beta \in \mathcal{D},\\
0, & \beta \notin \mathcal{D},
\end{cases}
\end{equation}
where $\xi_l^{-1}(\cdot)$ denotes the 3D IDWT. This construction restricts supervision to detail content while
producing a spatial-domain tensor, so the alignment loss is computed in the same geometry as standard
feature MSE (rather than directly on coefficient tensors whose layout depends on implementation choices). 


At each stage i, we align teacher and student projections as
\begin{equation}
\label{eq:stage_kd}
\mathcal{L}_{\mathrm{DCD}}^{(i)}
=
\frac{1}{N_i}\left\|
\mathcal{P}_{\mathcal{D}}\!\left(F_i^{(t)}\right)-
\mathcal{P}_{\mathcal{D}}\!\left(\phi_i\!\left(F_i^{(s)}\right)\right)
\right\|_2^2,
\end{equation}


The multi-stage distillation loss sums across stages:
\(
\mathcal{L}_{\mathrm{DCD}}
=
\sum_{i=1}^{M} \, \mathcal{L}_{\mathrm{DCD}}^{(i)}
\).

The overall training objective is:
\begin{equation}
\label{eq:total_loss}
\mathcal{L}_{\mathrm{total}} = \mathcal{L}_{\mathrm{seg}} + \mu\, \mathcal{L}_{\mathrm{DCD}},
\end{equation}
where $\mathcal{L}_{\mathrm{seg}}$ is the segmentation objective (Dice + cross-entropy) and $\mu$ balances distillation.
Since DWT/IDWT operators and $\phi_i$ are used only during training, DCD adds no inference-time overhead. Fig.~\ref{fig:overview}(b) shows the overall workflow of single-stage DCD distillation.

\section{Experiment}

\subsection{Dataset and Implementation}

\textbf{Datasets} We utilized the BraTS2024-BraTS-GLI dataset \cite{de20242024} and the ISLES 2022 dataset \cite{hernandez2022isles}. The BraTS 2024 dataset contains 1,350 cases (1,080 for training and 270 for validation). Each case includes four MRI modalities: pre-contrast T1-weighted (T1n), contrast-enhanced T1-weighted (T1c), T2-weighted (T2w), and T2-FLAIR (T2f). Tumor annotations are divided into four subregions: enhancing tissue (ET), non-enhancing tumor core (NETC), surrounding non enhancing FLAIR hyperintensity (SNFH), and resection cavity (RC). The ISLES 2022 dataset contains 250 cases (200 for training and 50 for validation), and each case includes three MRI modalities: apparent diffusion coefficient (ADC), diffusion-weighted imaging (DWI), and FLAIR. Its annotations consist of a single category: lesion.

\noindent \textbf{Implementation details.}
The proposed framework is implemented in PyTorch and trained on a single NVIDIA A100 GPU.
We adopt SGD with Nesterov momentum ($m=0.99$) using an initial learning rate of $0.01$ and weight decay of $3\times10^{-5}$.
We build the compact student using a channel reduction factor $r=4$. Both teacher and student follow the nnU-Net encoder and decoder topology. For Detail Consistent distillation, we use the Daubechies-4 wavelet (db4) as the wavelet basis, distilling the directional-detail subbands $\mathcal{D}$ at all encoder stages. The distillation weight is $0.05$.

\begin{table}[ht]
\centering
\caption{Performance comparison on BraTS 2024 and ISLES 2022.}
\label{tab:brats_isles}
\fontsize{8}{9}\selectfont
\setlength{\tabcolsep}{1.5pt}
\renewcommand{\arraystretch}{1.08}

\begin{tabular}{lccc ccc}
\toprule
\multirow{2}{*}{Method}
& \multicolumn{3}{c}{BraTS 2024 (Overall)}
& \multicolumn{3}{c}{ISLES 2022 (Overall)} \\
\cmidrule(lr){2-4}\cmidrule(lr){5-7}
& mDice(\%)$\uparrow$ & HD95(mm)$\downarrow$ & NSD(\%)$\uparrow$
& mDice(\%)$\uparrow$ & HD95(mm)$\downarrow$ & NSD(\%)$\uparrow$ \\
\midrule

\textcolor{teacherGray}{Teacher}
& \textcolor{teacherGray}{75.23\pmg{0.90}}
& \textcolor{teacherGray}{5.33\pmg{0.38}}
& \textcolor{teacherGray}{87.43\pmg{0.77}}
& \textcolor{teacherGray}{76.66\pmg{2.37}}
& \textcolor{teacherGray}{9.04\pmg{1.84}}
& \textcolor{teacherGray}{83.82\pmg{2.80}} \\

w/o KD
& 63.60\pmg{0.93}
& 7.84\pmg{0.46}
& 76.78\pmg{0.83}
& 70.21\pmg{3.03} 
& 18.94\pmg{3.82} 
& 81.22\pmg{2.65}  \\

CWD \cite{shu2021channel}
& 63.13\pmg{0.94}
& 8.29\pmg{0.52}
& 75.54\pmg{0.86}
& 71.94\pmg{2.95}
& 9.48\pmg{2.02}
& 82.74\pmg{2.67} \\

IFVD \cite{wang2020intra}
& 66.54\pmg{0.94}
& 7.09\pmg{0.45}
& 79.41\pmg{0.86}
& 72.43\pmg{2.98}
& 11.90\pmg{2.51}
& 83.30\pmg{2.69} \\

Logits \cite{hinton2015distilling} 
& 65.31\pmg{0.90}
& 8.43\pmg{0.51}
& 78.62\pmg{0.80}
& 71.51\pmg{3.03}
& 12.75\pmg{2.82}
& 82.66\pmg{2.72} \\

Feature \cite{Romero2015FitNets}
& 64.39\pmg{0.96}
& 7.59\pmg{0.48}
& 77.10\pmg{0.87}
& 72.70\pmg{2.90}
& 10.45\pmg{3.14}
& 83.58\pmg{2.67} \\

FreeKD \cite{Zhang2024FreeKD}
& 62.16\pmg{0.96}
& 8.91\pmg{0.53}
& 74.75\pmg{0.88}
& 71.03\pmg{3.05}
& 12.36\pmg{2.49}
& 81.86\pmg{2.72} \\

\midrule
\rowcolor{oursBlue}
Ours
& \textbf{68.51\pmg{0.92}}
& \textbf{6.25\pmg{0.36}}
& \textbf{81.34\pmg{0.82}}
& \textbf{73.95\pmg{2.70}}
& \textbf{12.95\pmg{2.49}}
& \textbf{83.47\pmg{2.78}} \\
\bottomrule
\end{tabular}
\end{table}

\subsection{Experiment Results}

\noindent \textbf{Comparative Results.} All results reported in the tables are presented as mean $\pm$ standard error of the mean. Table~\ref{tab:brats_isles} demonstrates that the proposed DCD surpasses existing distillation baselines on both BraTS 2024 and ISLES 2022.
On BraTS 2024, DCD improves the compact student from $63.60\%$ to $68.51\%$ mDice $+4.91\%$ (paired \(t\)-test \(p=8.64\times10^{-13}\), Wilcoxon signed-rank test \(p=4.40\times10^{-16}\)).
Compared with the previous best method (e.g., IFVD with $66.54\%$ mDice), DCD improves mDice by $+1.97\%$ (paired \(t\)-test \(p=0.0078\), Wilcoxon signed-rank test \(p=0.0080\)). On ISLES 2022, DCD achieves the highest $73.95\%$ mDice, improving over the non-distilled student by $+3.74\%$ (paired \(t\)-test \(p=0.0058\), Wilcoxon signed-rank test \(p=0.0008\)). Table~\ref{tab:brats_all} further suggests that DCD yields larger mDice gains on detail-sensitive regions. For example, NETC improves from $41.26\%$ to $54.36\%$ mDice $+13.10\%$.

\begin{table}[ht]
\centering
\caption{BraTS 2024 mDice results on NETC/SNFH and ET/RC.}
\label{tab:brats_all}
\fontsize{8}{9}\selectfont 
\setlength{\tabcolsep}{2pt}
\renewcommand{\arraystretch}{1.08}

\begin{tabular}{lcccc}
\toprule
Method & NETC mDice(\%)$\uparrow$ & SNFH mDice(\%)$\uparrow$ & ET mDice$(\%)\uparrow$ & RC mDice(\%)$\uparrow$ \\
\midrule
\textcolor{teacherGray}{Teacher} &
\textcolor{teacherGray}{60.85\pmg{3.20}} &
\textcolor{teacherGray}{90.17\pmg{0.59}} &
\textcolor{teacherGray}{76.18\pmg{2.08}} &
\textcolor{teacherGray}{73.75\pmg{2.07}} \\
w/o KD  & 41.26\pmg{2.81} & 87.00\pmg{0.67} & 61.57\pmg{2.37} & 64.61\pmg{2.17} \\
CWD \cite{shu2021channel}     & 40.97\pmg{2.77} & 87.26\pmg{0.69} & 59.91\pmg{2.38} & 64.39\pmg{2.17} \\
IFVD \cite{wang2020intra}    & 46.34\pmg{2.90} & 88.23\pmg{0.64} & 63.95\pmg{2.34} & 67.67\pmg{2.14} \\
Logits \cite{hinton2015distilling}  & 51.96\pmg{3.04} & 86.91\pmg{0.71} & 57.97\pmg{2.36} & 64.41\pmg{2.17} \\
Feature \cite{Romero2015FitNets} & 42.55\pmg{2.82} & 87.80\pmg{0.66} & 60.36\pmg{2.39} & 66.85\pmg{2.11} \\
FreeKD \cite{Zhang2024FreeKD}  & 39.75\pmg{2.75} & 86.39\pmg{0.70} & 58.71\pmg{2.38} & 63.80\pmg{2.17} \\
\midrule
\rowcolor{oursBlue}
Ours & \textbf{54.36\pmg{3.20}} & \textbf{88.49\pmg{0.63}} & \textbf{62.55\pmg{2.36}} & \textbf{68.62\pmg{2.12}} \\
\bottomrule
\end{tabular}
\end{table}

\begin{table}[ht]
\centering
\caption{Model complexity comparison on BraTS 2024 and ISLES 2022.}
\label{tab:complexity}
\fontsize{8.0}{9}\selectfont
\setlength{\tabcolsep}{4pt}
\renewcommand{\arraystretch}{1.08}

\begin{tabular}{lcc cc}
\toprule
\multirow{2}{*}{Model}
& \multicolumn{2}{c}{BraTS 2024}
& \multicolumn{2}{c}{ISLES 2022} \\
\cmidrule(lr){2-3}\cmidrule(lr){4-5}
& Params (M)$\downarrow$ & FLOPs (TF)$\downarrow$
& Params (M)$\downarrow$ & FLOPs (TF)$\downarrow$ \\
\midrule

\textcolor{teacherGray}{Teacher}
& \textcolor{teacherGray}{101.945}
& \textcolor{teacherGray}{19.505}
& \textcolor{teacherGray}{102.351}
& \textcolor{teacherGray}{17.707} \\

\rowcolor{oursBlue}
Student
& \textbf{6.381}
& \textbf{1.277}
& \textbf{6.406}
& \textbf{1.149} \\

\bottomrule
\end{tabular}
\end{table}

\begin{figure}[t]
  \centering
  \includegraphics[width=\linewidth]{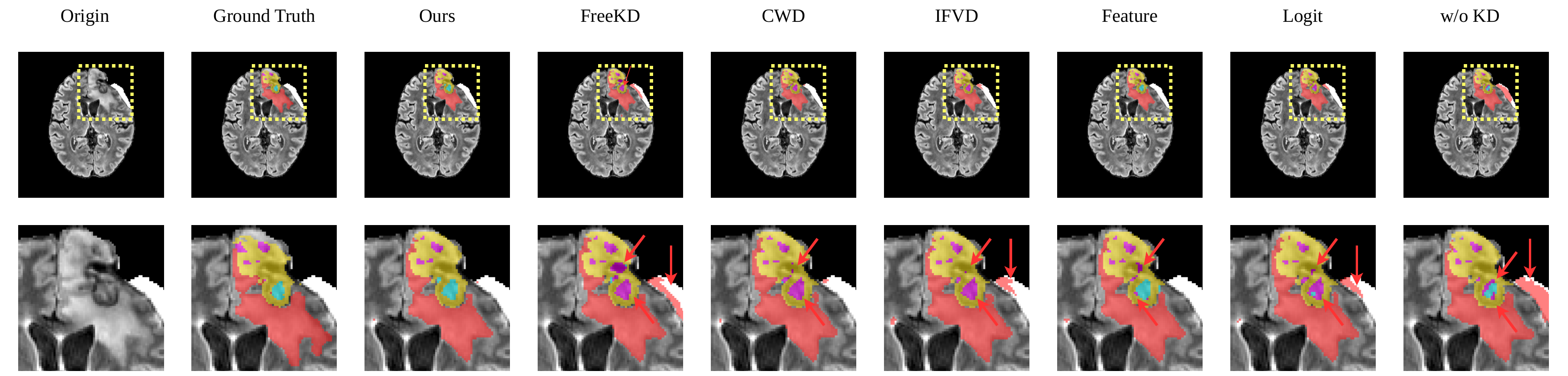}
  \caption{Qualitative comparison on BraTS 2024 dataset. The top row shows the segmentation predictions from different methods, and the bottom row provides zoomed-in views of the regions highlighted by the yellow dashed boxes. Compared with other methods, which exhibit mis-segmentation (red arrows highlight errors), DCD produces predictions closer to the ground truth.}
  \label{fig:QA}
\end{figure}

FreeKD, originally designed for natural images, yields limited gains for MRI segmentation, whereas DCD consistently improves mDice on both datasets. Table~\ref{tab:complexity} shows that DCD uses far fewer parameters and FLOPs than the teacher while preserving competitive accuracy. Fig.~\ref{fig:QA} further shows that DCD produces predictions better aligned with the ground truth.




\begin{table}[ht]
\centering
\caption{Ablation study on BraTS 2024 and ISLES 2022.}
\label{tab:ablation_brats_isles}
\fontsize{8}{9}\selectfont
\setlength{\tabcolsep}{1.5pt}
\renewcommand{\arraystretch}{1.08}

\begin{tabular}{lccc ccc}
\toprule
\multirow{2}{*}{Method}
& \multicolumn{3}{c}{BraTS 2024 (Overall)}
& \multicolumn{3}{c}{ISLES 2022 (Overall)} \\
\cmidrule(lr){2-4}\cmidrule(lr){5-7}
& mDice(\%)$\uparrow$ & HD95(mm)$\downarrow$ & NSD(\%)$\uparrow$
& mDice(\%)$\uparrow$ & HD95(mm)$\downarrow$ & NSD(\%)$\uparrow$ \\
\midrule

w/o KD
& 63.60\pmg{0.93} & 7.84\pmg{0.46} & 76.78\pmg{0.83}
& 70.21\pmg{3.03} & 18.94\pmg{3.82} & 81.22\pmg{2.65} \\
\addlinespace[2pt]

\multicolumn{7}{l}{\textbf{Band selection}} \\
\quad Band $\mathcal{S}$
& 61.98\pmg{0.96} & 8.91\pmg{0.53} & 74.80\pmg{0.90}
& 72.12\pmg{2.96} & 10.40\pmg{2.12} & 83.26\pmg{2.62} \\
\quad Band $\mathcal{A}$
& 62.25\pmg{0.96} & 8.75\pmg{0.51} & 74.64\pmg{0.90}
& 69.54\pmg{2.94} & 18.58\pmg{3.41} & 80.15\pmg{2.67} \\

\rowcolor{oursBlue}
\quad Band $\mathcal{D}$
& \textbf{68.50\pmg{0.92}} & \textbf{6.25\pmg{0.36}} & \textbf{81.34\pmg{0.82}}
& \textbf{73.95\pmg{2.70}}
& \textbf{12.95\pmg{2.49}}
& \textbf{83.47\pmg{2.78}} \\
\addlinespace[2pt]

\multicolumn{7}{l}{\textbf{IDWT}} \\
\quad w/o IDWT
& 62.87\pmg{0.91} & 9.01\pmg{0.51} & 77.13\pmg{0.84}
& 70.06\pmg{3.11} & 15.69\pmg{2.79} & 80.51\pmg{2.80} \\

\rowcolor{oursBlue}
\quad w/ IDWT
& \textbf{68.51\pmg{0.92}} & \textbf{6.25\pmg{0.36}} & \textbf{81.34\pmg{0.82}}
& \textbf{73.95\pmg{2.70}}
& \textbf{12.95\pmg{2.49}}
& \textbf{83.47\pmg{2.78}} \\
\bottomrule
\end{tabular}
\end{table}

\noindent \textbf{Ablation Results.}
Table~\ref{tab:ablation_brats_isles} reports ablation studies validating the contribution of each component.
\textbf{(i) Subband selection.}
Distilling $\mathcal{D}$ yields the best results, while distilling $\mathcal{S}$, i.e., the extreme high frequency band, performs worse than w/o KD on BraTS 2024 ($61.98\%$ vs. $63.60\%$ mDice). This supports our motivation that the noisiest high-frequency band can provide harmful supervision. In contrast, $\mathcal{D}$ achieves $68.50\%$ and $73.95\%$ mDice on BraTS 2024 and ISLES 2022, outperforming both $\mathcal{S}$ and $\mathcal{A}$.
\textbf{(ii) IDWT reconstruction.}
Removing IDWT leads to a marked degradation on both datasets, with BraTS 2024 mDice dropping from $68.51\%$ to $62.87\%$ and HD95 increasing from $6.25$ to $9.01$, and ISLES 2022 mDice decreasing from $73.95\%$ to $70.06\%$.




\section{Conclusion}
We propose Detail Consistent Distillation (DCD), a stage-wise distillation framework for efficient 3D medical image segmentation. DCD performs detail subspace selection at each encoder stage and distills detail subbands, enabling the student to focus on fine-grained, task-relevant cues without over-regularizing global semantics and amplifying noise. Extensive experiments on BraTS 2024 and ISLES 2022 demonstrate that DCD is robust and efficient for MRI segmentation.

\begin{credits}
\subsubsection{\ackname}
The authors gratefully acknowledge the computational resources provided by the Delta system at the National Center for Supercomputing Applications (NCSA) [award OAC 2005572] through ACCESS allocations CIS260331 and CIS250976. ACCESS, the Advanced Cyberinfrastructure Coordination Ecosystem: Services \& Support program, is supported by U.S. National Science Foundation (NSF) grants \#2138259, \#2138286, \#2138307, \#2137603, and \#2138296. This work was partially supported by NIH grant U01AG079847.

\subsubsection{\discintname}
The authors have no competing interests to declare that are relevant to the content of this article.

\end{credits}
%
%
%
\FloatBarrier
\bibliographystyle{splncs04}
\bibliography{mybibliography}
%




\end{document}